\documentclass{article}
\usepackage{ijcai21}

\usepackage{times}

\usepackage{url}
\usepackage[hidelinks]{hyperref}
\usepackage[utf8]{inputenc}
\usepackage[small]{caption}
\usepackage{graphicx}
\usepackage{amsmath}
\usepackage{booktabs}
\usepackage{tabularx}
\usepackage{makecell}
\usepackage{soul}
\usepackage{color,soul}

\usepackage{algorithm}
\usepackage[noend]{algpseudocode}

\usepackage{xspace}
\newcommand{\eg}{EG\xspace}
\newcommand{\etg}{ETG\xspace}
\newcommand{\hierarchy}{\ensuremath{\mathcal{H}}\xspace}
\DeclareMathOperator{\coo}{coord}

\newtheorem{example}{Example}

\title{Streaming and Learning the Personal Context}

\author{
Fausto Giunchiglia$^1$\footnote{Contact Author}\and
Marcelo Rodas Britez$^1$\and
Andrea Bontempelli$^1$\And
Xiaoyue Li$^1$
\\
\affiliations
$^1$University of Trento, Trento, Italy\\
\emails
\{fausto.giunchiglia, marcelo.rodasbritez, andrea.bontempelli, xiaoyue.li\}@unitn.it
}

\begin{document}

\maketitle  

\begin{abstract}
The representation of the personal context is complex and essential to improve the help machines can give to humans for making sense of the world, and the help humans can give to machines to improve their efficiency.
We aim to design a novel model representation of the personal context and design a learning process for better integration with machine learning. We aim to implement these elements into a modern system architecture focus in real-life environments. 
Also, we show how our proposal can improve in specifically related work papers.
Finally, we are moving forward with a better personal context representation with an improved model, the implementation of the learning process, and the architectural design of these components.
\end{abstract}

\section{Introduction}
\label{section:introduction}

Every person makes sense of their personal context differently because of their different sets of personal characteristics (intelligence) and behaviour (life choices). However, the machine's understanding of the personal context is radically different from the user's understanding. This limitation is due to the limited definition of the personal context, and the lack of tools to make sense of the personal context. For instance, while the person you are with now can be linked to a name, for people it has more meaning than just a name, e.g., friend and colleague. Additionally, these meanings are not fixed, they may change at any time, and every person can assign additional meaning using different criteria. Thus, effective context recognition requires a complex and dynamic representation of the personal context and the collaboration of the people to fill the cognitive gap of machines.

The addition of human collaboration into the context recognition learning of machines is an important part of supervised machine learning \cite{vapnik2013nature}. These interactions bring new challenges to the implementation of machine learning algorithms. For instance, humans can be defined as the expert of the supervised algorithms, interacting in an offline fashion by annotating sensor data \cite{webb2003statistical}, or the interaction can be directly online, as active learning \cite{settles2009active,hoque2012aalo,hossain2017active}. The human collaboration is important when we are moving into real-life scenarios \cite{kwapisz2011activity}.

Other challenges of this collaborative approach are the possibility of overwhelming the humans and the possible differences between the assignment of meaning between people \cite{chang2017investigation}, thus, making the annotation a personal activity. Then, having humans as the input of the information opens the possibility of human error in the collaboration process \cite{tourangeau2000psychology}, and this issue is well known in social sciences and psychology, because of response biases in answering self-reports \cite{west2013quality}, and more importantly, these biases are not well-understood \cite{freedman2013interviewer}.

We propose a novel context model based on the work from  \cite{giunchiglia2018personal}. That work focused on ensuring the reliability of annotations, whereas our focus is on improving personal context representations to get closer to work in real-life scenarios. So, we propose to add a more precise representation of personal context that can also work with machine learning algorithms. We formalize a context model based on ontology and use it with the streaming data to have a knowledge representation of context data. This formalization allows moving towards a generic definition of context that can work with existing multi-label machine learning approaches, using a conversion algorithm. Eventually, the last piece of the puzzle will be the design and development of the Streaming System to manage technically the dynamic context data, and it will be organized in the system architecture with modular components for independent development and easy deployment in current cloud environments.

Some examples of our model improvement can be seen compared with our main related work \cite{giunchiglia2018personal,bontempelli2020learning,zeni2019fixing}. All of them can take benefit from our novel personal context representation and can use our conversions algorithms to explicitly implement the transformations needed for the machine learning algorithms.

The paper is structured as follows. Section~\ref{section:TheContextInTime} introduces context modeling. Section~\ref{section:TheCurrentContext} illustrates our representation of personal context, while we provide the formal representation in Section~\ref{section:TheCurrentContextKnowledgeGraph}. Then, we show the learning process to transition from our formal representation to machine learning representation in Section~\ref{section:learningcontext} and how our formal representation can be converted to a Direct Acyclic Graph (DAG). Finally, Section~\ref{section:relatedwork2} describes works related to ours, and Section~\ref{section:conclusion} concludes our paper.

\section{The context in time}
\label{section:TheContextInTime}

When we talk about the context, we concentrate on the context of a person called observer. The observer's context is the representation of a partial view of the world. We describe this context into three main dimensions: the viewpoint, the part-whole relation, and the endurant-perdurant.

Firstly, we have the viewpoint dimension divided as outside viewpoint and observer viewpoint. The outside viewpoint is the view of an ideal observer who can describe everything from a certain point of view. We distinguish this view from the world's static and dynamic properties. In the static property, there are whatever does not change in time, e.g., mountains, buildings, streets. In the dynamic property, there are not only the moving people, but also the moving animals, and facilities in their manifestations, like trains. Then, the observer viewpoint describes how the observer perceives what is around her or him. In this view, we also have the property of being static or dynamic, but it is relative to the movements of the observer.

Secondly, context is a part-whole relation. In our everyday life, when we do things, we are always embedded in the world. From an ontological view, we are part of the whole world. Thus, we call reference context as the element of the outside context with a volume and extension that is large enough to contain all our movements and changes. For instance, the reference context is the city of Trento when the user walks around, or the users' home when they are at home. In turn, our body as a whole has parts (e.g., arms, mind, legs) that are with us all the time, and they define the internal context of the user. The internal context identifies the elements of the user's body at different levels of abstraction. We usually distinguish between physical parts, such as arms, body, fingers, and mental parts, such as mind, memory, emotions. The context as a part-whole relation is divided into reference context and internal context, and both contexts with different dimensions play a role in our life.

Thirdly, context as endurant and perdurant refers to the bigger relation to changes in time or space. Events and actions are perdurants and elements, like $me$, are endurants.

\subsection{The spatio-temporal context}

The context, as viewpoints, defines the reference point from which we construct the context and the context, as part-whole, defines which parts we should consider. So, next, we need to define how we keep track of the context from a quantitative point of view, with a set of quantitative and qualitative measures. Therefore, based on these measures, we introduce the spatio-temporal context.

The spatio-temporal context consists of the temporal and spatial reference context. The former includes dates, times and all the additional notations like weekdays and seasons. The latter contains the world coordinate system. There are various reasons why context should be represented as a spatio-temporal context. First, this is a common representation when we think of the world. Second, any device today can easily retrieve the time and time zones, and the space coordinates (e.g., via GPS). Third, time can be used to measure the changes in all the elements of the world, all evolving at different speeds, thus temporal context allows us to use them together based time. Finally, a lot of data about the spatial reference context and its sub-contexts are available from external sources (e.g., Google Maps, OpenStreetMaps).

The spatio-temporal context, also called objective context, at time $t$ is defined as:
\begin{align*}
    o_t = (D_t, T_t:\; &L_t,\ me,\ \coo_t(me),\\
            &P^1_t: \coo_t(P^1),\dots, P^k_t: \coo_t(P^k),\\
            &O^1_t: \coo_t(O^1),\dots, O^m_t: \coo_t(O^m))
\end{align*}
where $D_n$, $T_n$ stands for date and time respectively, $L_n$ is the location, namely the smallest possible spatial reference context that we can compute. Here $me$ is the observer, $P^i$ are persons and $O^m$ are objects. The function $\coo(\dots)$ computes the spatial coordinates of $me$, objects, and persons.

The number and type of persons and objects change over time. Hence, we will have a sequence of time-tagged states, namely $\mathcal{O} = \{o_1,\dots,o_n\}$. We call the sequence $\mathcal{O}$ as the streaming context. In the streaming context, within the given reference location, it is easy to compute spatial relations (e.g., near, right, left, in front, far relative to the location) of the different elements among themselves. For instance, the system can compute that the smartphone is \textit{in} the home building and the smartphone is \textit{near} the computer.

\subsection{The objective and subjective context}

The spatio-temporal context is also called objective context, since all the relations are computed concerning what is objectively measured, in terms of spatial relations. However, notice that different observers will have different views of the world. For instance, the school building has the function of study-place from the point of view of a student and is the teacher's workplace. The word function here is used with the precise meaning defined in \cite{giunchiglia2017teleologies,giunchiglia2018personal}. Hence, ``\textit{the function of an object formalizes the behavior that an object is expected to have}'' \cite{giunchiglia2017teleologies}. For instance, objects are trains and buildings. The expected behavior may be the purpose of the object (e.g., fridge) or the role of a person (e.g., friend).

The subjective context includes both the objective context elements and the function of persons and objects as seen by the user. Thus, the subjective context at time $t$ is defined as:
\begin{align*}
s_t = &(D_t, T_t:\; L_t, me, \coo_t(me),  F_t(P^1), \dots, F_t(P^k), \\ 
    &F_t(O^1), \dots, F_t(O^m)),
\end{align*}
where $F_n(P^k)$ and $F_n(O^m)$ are the functions with respect to a person $P^k$ and an object $O^m$, respectively. The number and type of persons and objects, and their functions change over time. The sequence of subjective contexts over time is defined as the subjective streaming context $\mathcal{S} = \left\{ s_1,\dots,s_n\right\}$.

\subsection{The endurant and perdurant context}

In the endurant context \cite{giunchiglia2017personal}, its parts are endurants, essentially objects where the spatial extension of their actions is contained by the space defined by the spatial (reference) context. The actions ``\textit{represents how objects change in time}'' \cite{giunchiglia2017teleologies}. For instance, running in a park performed by a runner. We also need to represent actions, in particular, the actions that are executed by the endurant $me$ and also by any other elements of the outside dynamic context. Actions can be seen in two ways: \textit{(i)} actions modeled as processes, namely as sequences of single micro-steps, each of length close to zero; \textit{(ii)} actions modeled as events, which are often also called perdurants, namely as complete movements which last for a certain duration. Actions as events have key properties, similar to endurants. An event and an action can associate with a set of component sub-events and sub-actions. 

Considering the mentioned concepts, the fundamentally different role of space and time should become clear. Whereas the parts of the space context are only used to limit the space where things happen, the parts of time have the main goal to detail how actions get executed. Things get complicated by adding objects, people, and functions as shown in the data representations shown in Table~\ref{fig:streamingContext}.

It is worth noting that each function of a person is associated with a limited set of actions, and the type of action that a person can perform can be considerably limited by knowing his or her function.

The actions apply to $me$ and persons, whereas functions apply to $me$, other objects and persons. Notice also, that the stored location $L$ is limited by the most specific location that we can compute. This is because the bigger locations are assumed to be static and stored in the system. For events, instead, we store the smallest possible most general event as well as those component actions which are done during a certain period. Thus, for instance, the action/event meeting can have sub-actions such as talking, walking, listening, typing. 
Table~\ref{fig:StreamContextMatrix} reports the streaming context matrix of Example~\ref{ex:travel} and shows how the context changes over time.

\begin{table*}[tb]
\begin{align*}
\{\\ &\left(D_1, T_1: \mathrm{super}(L_1), \mathrm{super}(E_1), L_1, E_1, me, \coo_1(me), A_1^{me}, F_1(P^1):A_1^{P^1}, \ldots, F_1(P^k):A_1^{P^k}, F_1(O^1), \ldots, F_1(O^m)\right),\\
&\left(D_2, T_2: \mathrm{super}(L_2), \mathrm{super}(E_2), L_2, E_2, me, \coo_2(me), A_2^{me}, F_2(P^1):A_2^{P^1}, \ldots, F_2(P^k):A_2^{P^k}, F_2(O^1), \ldots,  F_2(O^m)\right),\\
& \dots,\\
&\left(D_n, T_n: \mathrm{super}(L_n), \mathrm{super}(E_n), L_n, E_n, me, \coo_n(me), A_n^{me}, F_n(P^1):A_n^{P^1}, \ldots, F_n(P^k):A_n^{P^k}, F_n(O^1), \ldots, F_n(O^m)\right)&\\
\}
\end{align*}
\caption{The personal streaming context, where $E_n$ is an event, $\mathrm{super}(L_n)$ and $\mathrm{super}(E_n)$ are the super-classes of $L_n$ or $E_n$, respectively. The set of actions performed by \textit{me} or by the persons based on their functions are denoted with $A_n^{k} = \{a_1,\ldots, a_i\}$, with $k \in \{me, P^1, \dots, P^k\}$.}
\label{fig:streamingContext}
\end{table*}

\begin{table*}[ht]
    \centering
    \setlength{\tabcolsep}{4pt} 
    \small
    \begin{tabular}{|c|c|c|c|c|c|c|p{1.2cm}|c|c|}
         \hline
         $D_n$ & $T_n$ & $\mathrm{super}(L_n)$ & $\mathrm{super}(E_n)$ & $L_n$ & $E_n$ & $\coo_n(me)$ & $A_n^{me}$ & $F_n(P^1):A_n^{P^1}$ & $F_n(O^1)$ \\
         \hline
        02/06/2021 & 12:15 & Trentino & Travel 1 & Train 1 & Take\ Train & x41, y41, z41 & Sitting & \textit{NaN} &  \makecell{RestToolOf(\\Xiaoyue, Seat 1)}\\
        \hline
        02/06/2021 & 12:30 & Trentino & Travel 1 & Roads 2 & Walk &  x43, y43, z43 & Walking, Talking  & \makecell{FriendOf(\\Xiaoyue, Haonan):\\Walking, Listening} & \textit{NaN}\\
        \hline
    \end{tabular}
    \caption{A streaming context matrix representing the travel scenario of Example~\ref{ex:travel} from the point of view of Xiaoyue, i.e., the observer $me$. $P^1$ is Haonan and $O^1$ is the object ``Seat 1''. Each column is a property, and every row stands for the current context in a specific timestamp.}
    \label{fig:StreamContextMatrix}
\end{table*}

\section{The current context}
\label{section:TheCurrentContext}

\begin{figure*}[!htb]
    \centering
    \includegraphics[width=\textwidth]{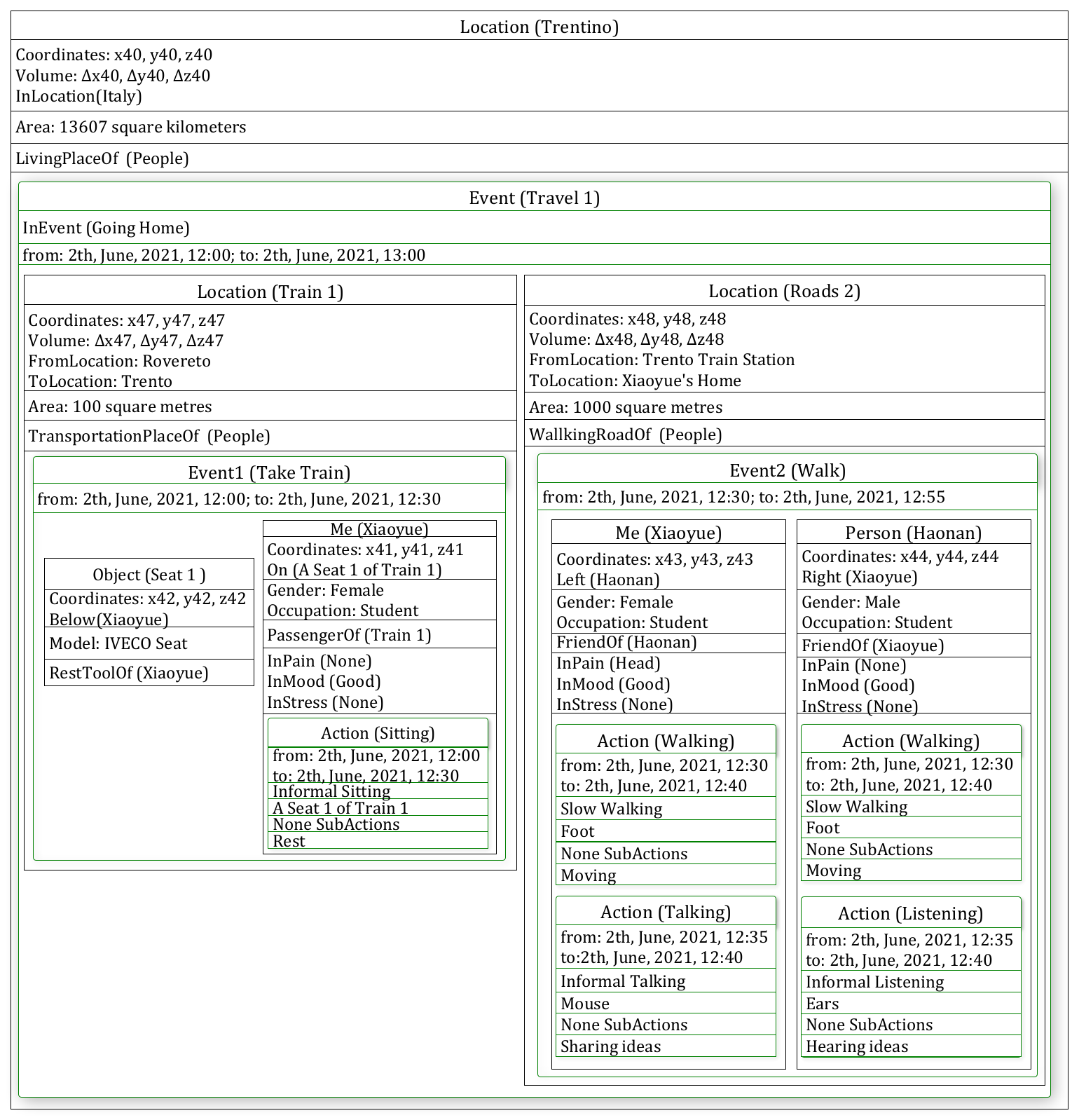}
    \caption{One Event Multiple Locations: a travel around $me$. Representation of the Example~\ref{ex:travel}.}
    \label{fig:dimensionsofcontext}
\end{figure*}

The streaming context describes the contexts of an observer according to time. To represent each context occurrence, we define a set of notions to build a figure of the current context. We mainly divide the current context into four types of context according to how things compose in space and time. From top to bottom we have the following cases:
\begin{itemize}
    \item 1L1E (One Location One Event), such as a lecture holds in a classroom;
    \item 1LME (One Location Multiple Events), such as a sequence of meetings hold in an office, or eat breakfast, lunch and dinner at home; 
    \item 1E1L (One event One Location), this is the same as 1L1E. The difference is in the focus, which is now the location, e.g., the classroom where the lecture holds;
    \item 1EML (One Event Multiple Locations), such as a travel goes from many different places;
\end{itemize}
An 1EML example is shown in Figure~\ref{fig:dimensionsofcontext}, which describes the following travel scenario around observer $me$.

\begin{example}\label{ex:travel}
In the travel scenario, $me$ is Xiaoyue, she has travel named Travel 1 in the Trentino of Italy from 12:00 to 13:00 of 2th, June 2021. From 12:00 to 12:30 on this day, she takes train 1 from Rovereto to Trento, she sits on seat 1 by herself. From 12:30 to 12:55 on the same day, she walks on Roads 2 from Trento Train Station to Xiaoyue's Home, together with her friend named Haonan. In addition, Xiaoyue talks to Haonan, and Haonan listens to Xiaoyue when they are walking. This scenario involves one event travel and multiple locations.
\end{example}

In general, in the Figure~\ref{fig:dimensionsofcontext}, all elements are divided into Perdurants and Endurants. Perdurants are the Event and Action, and Endurants include Person, Object, and Location. An Event happens in a Location, a Person and Object appear in an Event and Action is in a Person. Most elements' inclusion relationships $IN$ can be represented by the positions of those elements' internal boxes. For the top-level Location and Event, we can add an extra attribute box for them respectively as $InLocation(Location)$ and $InEvent(Event)$, to represent their belongs.

In the rest of this section, we list the attributes of Location, Event, Object, Person and Action, each kind of attributes is represented as a box in the figure. The \textbf{Location} has the following attributes:
\begin{itemize}
    \item Spatial properties: Coordinates ($x_i$, $y_i$, $z_i$), Volume ($\Delta x_i$, $\Delta y_i$, $\Delta y_i$) , and InLocation($L_i$) shows the super Location for the top level Location;
    \item Visual properties, namely some properties of Location that can be observed visually;
    \item Location' functions: $FunctionOf(U)$ with $U \in \{P^1,\dots,P^k, O^1,\dots,O^M\}$, which shows location's functions for persons and objects;
    \item An extra box: including the rest part of the context that happens in the Location.
\end{itemize}
The \textbf{Event} orders by time and is represented by a box with round corners. Events include actual events and virtual events. An Event can have Sub-Events, the Event and Sub-Event have the following attributes:
\begin{itemize}
    \item Super Event for the top level event: $InEvent(E_i)$, which shows the super event $E_i$ for the current Event;
    \item Temporal properties: Begin Time - End Time ($Date_i$, $Time_i$ - $Date_j$, $Time_j$), which shows the date and time of the start and end of the event;
    \item An extra box: including the rest part of the context that happens in the Event.
\end{itemize}
The \textbf{Object} appears in Event and has the following attributes:
\begin{itemize}
    \item Spatial properties: Coordinates ($x_i$, $y_i$, $z_i$), In/Far/... ($P^i$/$O^m$/...);
    \item Visual properties, namely some properties of Object that can be observed visually;
    \item Object's functions: $FunctionOf(U)$ with $U \in \{P^1,\dots,P^k, O^1,\dots,O^M\}$, which shows Object's functions for persons and other objects.
\end{itemize}
The \textbf{Person} appears in Event and has the following attributes:
\begin{itemize}
    \item Spatial properties: Coordinates ($x_i$, $y_i$, $z_i$), In/Far/...($P^i$/$O^m$/...);
    \item Visual properties, namely some properties of Person that can be observed visually;
    \item Person's functions: $FunctionOf(U)$ with $U \in \{P^1,\dots,P^k, O^1,\dots,O^M\}$, which shows Person's functions for other persons and objects; 
    \item Internal states: Physical states ($InPain()$), Mental states ($InMood()$, $InStress()$);
    \item An extra box: including the Actions of Person.
\end{itemize}
The \textbf{Action} is similar with Event, it orders by time and is represented by a box with round corners. An Action can has many Sub-Action, the Action and Sub-Action have following attributes, each attribute has a box for itself.
\begin{itemize}
    \item Temporal properties: Begin Time - End Time ($Date_i$, $Time_i$ - $Date_j$, $Time_j$), which shows the date and time of the start and end of the Action;
    \item Visual properties, namely some properties of Action that can be observed visually;
    \item Means of the Action: Means($O^m$/...);
    \item Sub-action: Action $i$;
    \item Action's functions: $FunctionOf(U)$ with $U \in \{P^1,\dots,P^k, O^1,\dots,O^M\}$, which shows Action's functions for persons and objects.
\end{itemize}

\section{The current context as a Knowledge Graph}
\label{section:TheCurrentContextKnowledgeGraph}

We use the \emph{Entity Type Graph} (ETG) and the \emph{Entity Graph} (EG) in ontology to represent the context. ETG is a knowledge graph where nodes are entity types, which are further decorated with data properties. The object properties are presented in the graph representing the relations among the entity types.

\begin{figure*}[htb]
    \centering
\includegraphics[width=\textwidth]{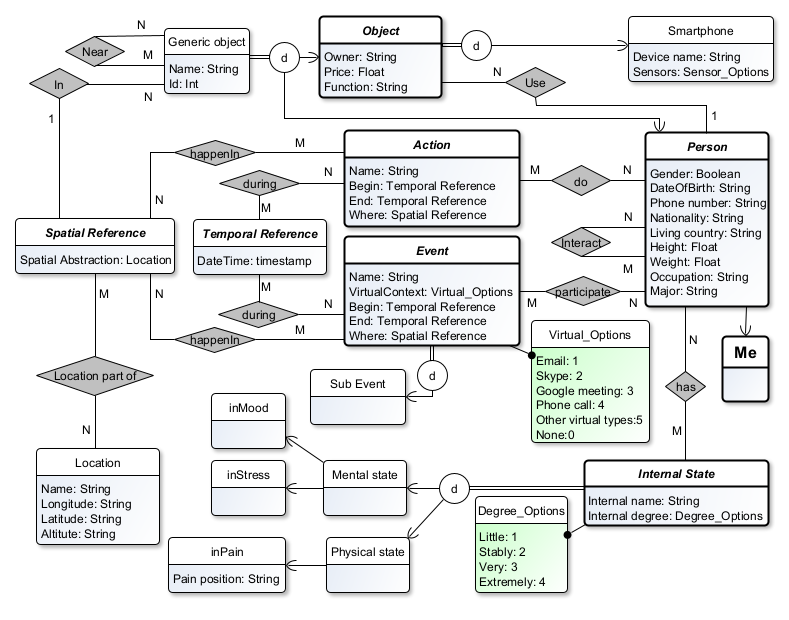}
   \caption{An ETG representing partially the personal context in our travel example.}
    \label{ETG2-1}
\end{figure*}

In Figure~\ref{ETG2-1}, white box nodes are entity types that include data properties with data types, the green box nodes enumerates the values of data property. The object properties are connecting all entity types, represented by diamond nodes with arrows. The inheritance relation in the ETG is represented by an arrow from the super-class to the sub-class, and the sub-class inherits all the data properties and object properties of its super-class.
The EG populates the entity types and properties defined in the ETG with specific values. It is a data graph where nodes are entities that are connected by object property values representing the relations. Each entity further includes data property values. The streaming context can be viewed as a stream of EGs, in which each EG describes the context at a different time. 

\begin{figure*}[htb]
    \centering
    \includegraphics[width=0.95\textwidth]{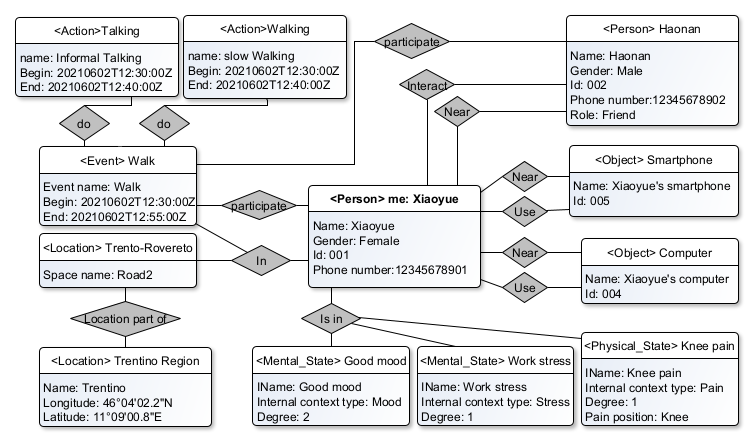}
    \caption{An EG representing partially the scenario about travelling of Example~\ref{ex:travel}.}
    \label{EG2-1}
\end{figure*}

We design an EG example as Figure~\ref{EG2-1} according to the scenario in Example~\ref{ex:travel}. The graph represents the context around "Me", this contains entities shown by nodes, e.g., "Smartphone", "Talking", "Walk". Also, we can see object property values.

\section{Learning Context}
\label{section:learningcontext}

AI applications like smart personal assistants provide a service to the users based on their context.
The context information is usually not available to the machine, and hence it has to infer the location or the activity of the user from sensor data (e.g., GPS, accelerometer, nearby Bluetooth devices). In our scenario, Xiaoyue is carrying a smartphone that generates a stream of sensor readings, and she annotates the data by answering questions about her context, e.g., ``Where are you'' and ``What are you doing?''. The sensor data are aggregated in time windows generating a stream of instances (e.g., the average number of nearby Bluetooth devices in the last 30 minutes). On each incoming instance, the machine decides whether to query the user to acquire the labels. The machine learning technique defines the query strategy, and in the simplest case, the labels are acquired on every instance. 

The user's context recognition is a supervised learning problem in which an instance $\mathbf{x}$ is associated to multiple concepts $\mathbf{y}$ (aka classes in machine learning). The concepts are organized in a ground-truth hierarchy $\hierarchy = (C, I)$, which is a direct acyclic graph (DAG) where nodes $C = \{1,\dots,c\}$ are the concepts and edges $I \subset C \times C$ are \textit{is-a} relations, i.e.,  $I = \{(c_i, c_j) \, | \, c_i, c_j \in C \, \textrm{and}\, c_i \, \textrm{is a child of}\, c_j\}$ \cite{silla2011survey}. The labels of the instances are indicator vectors $\mathbf{y} \in \{0,1\}^c$, where the $i$-th elements is 1 if $\mathbf{x}$ belong to $i$-th concept in \hierarchy and 0 otherwise. The machine is trained on a stream of examples $\mathbf{z}_t = (\mathbf{x}_t, \mathbf{y}_t)$ drawn from a ground-truth distribution $P(\mathbf{X}, \mathbf{Y})$ that is always consistent with a ground-truth hierarchy \hierarchy, i.e., if there is an edge from class $c_i$ to class $c_j$, then $y^i = 1$ implies $y^j = 1$ and conversely $y^j = 0$ implies $y^i = 0$. The goal of this hierarchical classification tasks is to learn a classifier that recognize well the context on future sensor readings.

The \etg and \eg introduced in Section~\ref{section:TheCurrentContextKnowledgeGraph} can be used as prior knowledge about the structure of the hierarchy. They encode the available information about the user and the world. Algorithm~\ref{alg:kr2ml} shows the conversion from ETG and EG to a DAG \hierarchy.
The first step is to convert each entity type (etype) in the ETG as a node in \hierarchy (lines \ref{kr2ml:start_etype} - \ref{kr2ml:end_etype}).
Second, each entity in EG also becomes a node that is added as a child of the node referring to the etype of the entity (lines \ref{kr2ml:start_instances} - \ref{kr2ml:end_instances}). The hierarchy encodes the information about the current user, so the \textit{Me} etype and the corresponding entity (e.g., Xiaoyue entity in Figure~\ref{EG2-1}) are not considered.

\begin{figure}
    \centering
    \includegraphics[scale=0.5]{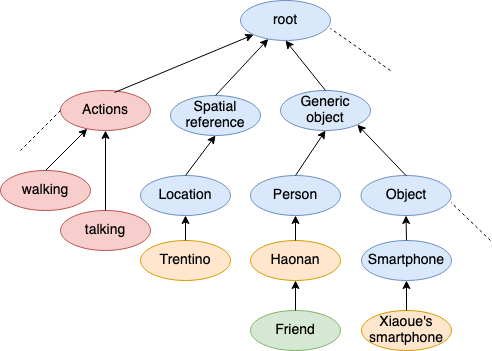}
    \caption{A partial representation of the DAG obtained from \etg and \eg examples in Section~\ref{section:TheCurrentContextKnowledgeGraph}. Orange nodes are derived from \eg. Blue nodes are derived from \etg. Red nodes are actions and green nodes are the functions.}
    \label{fig:kr2ml}
\end{figure}

The properties of the \etg are grouped in properties that are context depends and properties that are static. The value of the former changes every time the users change their context and, in Figure~\ref{ETG2-1} are $Q=$ \{\textit{near}, \textit{use}, \textit{interact}, \textit{in}, \textit{do}, \textit{happenIn}, \textit{during}, \textit{participate}\}. For instance, if Xiaoyue travel from the city of Trento to Rovereto, the \textit{in} property will change accordingly. In contrast, the fact that Trento is \textit{partOf} Italy can be assumed to remain valid even if user's context is changed. This distinction is necessary since the value of context-dependent properties are derived from the output of the context recognition task (e.g., the machine recognizes that the user is in the city of Trento and thus updates the \textit{in} property accordingly).
The object properties that are not contextual are converted as follow. Given a object property $p \in$ \{\textit{isA}, \textit{partOf}, \textit{has}\} that links the etype $A$ to $B$, then the node referring to etype $A$ becomes a child of the node of etype $B$. For the other object properties, a new node referring to the property is added as child of the codomain etype node (lines \ref{kr2ml:start_p_etype} - \ref{kr2ml:end_p_etype}). For every object property value $i$ of the property $p$ linking the entities $a$ and $b$, a new node $c_i$ is added as child of $c_p$ (i.e., the node referring to the property $P$), and as parent of $c_b$ (i.e., the node pointing to the entity $b$) (lines \ref{kr2ml:start_p_instances} - \ref{kr2ml:end_p_instances}).
Finally, all nodes that does not have a parent are connect to the root node and the transitive reduction is applied (lines \ref{kr2ml:start_root_node} - \ref{kr2ml:end_root_node}).  
Figure~\ref{fig:kr2ml} shows an extract of the DAG resulting from applying Algorithm~\ref{alg:kr2ml} on EG and ETG presented in Section~\ref{section:TheCurrentContextKnowledgeGraph}.

Every node in \hierarchy has a unique identifier that is used to reference back to the ETG and EG. The node name can be translated into a human-readable text that is used to interact with the user. This aspect is left as future work. The concept hierarchy available at the beginning can evolve over time and has to be continually updated. This aspect has been defined as knowledge drift and is addressed in~\cite{bontempelli2021human}.

\begin{algorithm}[!tb]
    \caption{Convert \etg and \eg in a DAG.}
    \label{alg:kr2ml}
    \begin{algorithmic}[1]
        \Statex \textbf{Inputs}: \etg, \eg, and the set of context dependent object properties $Q$
        \Statex \textbf{Outputs}: $\hierarchy = (C, I)$
        \Statex
    \State $C \gets \emptyset$
    \State $I \gets \emptyset$
	\For{every etype $A \ne $\xspace\textit{Me} in \etg} \label{kr2ml:start_etype}
	        \State let $c_A$ being the node of etype $A$
            \State $C = C \cup \{c_A\}$ 
	\EndFor \label{kr2ml:end_etype}
        \Statex
	\For{every entity $a$ such that $\neg$\xspace\text{Me}$(a)$ in \eg} \label{kr2ml:start_instances}
	        \State let $c_a$ being the node of the entity $a$
	        \State let $c_A$ being the node of the etype of $a$ 
            \State $C = C \cup \{c_a\}$
            \State $I = I \cup \{(c_a, c_A)\}$
	\EndFor \label{kr2ml:end_instances}
        \Statex
        \For{every object property $p$ in \etg such that $p \notin \mathcal{Q}$ }
            \State let $c_p$ being the node of $p$
	        \For{every $p(A, B)$ in \etg} \label{kr2ml:start_p_etype}
	            \State let $c_A$ and $c_B$ being the nodes of entity $A$ and $B$ respectively
                \If{$p \in$ \{\textit{isA}, \textit{partOf}, \textit{has}\}}
                    \State $I = I \cup \{(c_A, c_B)\}$
                \Else
                    \State $C = C \cup \{c_p\}$
                    \State $I = I \cup \{(c_p, c_B)\}$
                \EndIf
            \EndFor \label{kr2ml:end_p_etype}
	        \For{every $p(a,b)$ in \eg} \label{kr2ml:start_p_instances}
                \State let $c_i$ being the node encoding $p(a,b)$
                \State let $c_b$ being the node of entity $b$
                \State $C = C \cup \{c_i\}$
                \State $I = I \cup \{(c_i, c_p)\}$
                \State $I = I \cup \{(c_b, c_i)\}$
            \EndFor \label{kr2ml:end_p_instances}
        \EndFor
        \Statex
    \State let $c_0$ being the root node
	\State $C = C \cup \{c_0\}$ \label{kr2ml:start_root_node}
        \For{every node $c \in C$ that $\mathrm{parent}(c) = \emptyset$}
        	\State $I = I \cup \{(c, c_0)\}$ add an edge from $c$ to $c_0$
        \EndFor
	\State apply transitive reduction on $\hierarchy = (C, I)$ \label{kr2ml:end_root_node}
	
    \end{algorithmic}
\end{algorithm}

\section{Related Work}
\label{section:relatedwork2}

Considering our novel context modelling of the personal context, the context learning, and the architecture presented in this paper, we can show as use cases the papers that individually compartmentalizes examples for these improvements.
Current works on context recognition focus on learning the relationship between the input data (sensor readings) and the target concepts (context). The structure of the context is implicitly learned by the implemented machine learning algorithm during the training phase. The new parts are described in Section~\ref{section:TheCurrentContext} and compared with related work we can outline the following: 1) the connection between the learned context model; 2) The extension of new dimensions and classification of context, i.e. Internal State, Functions, Actions. The Algorithm~\ref{alg:kr2ml} bridges the gap between the two modellings allowing context recognition to leverage both machine learning solutions and knowledge graph representation (ETG and EG). For instance, the ETG and EG can generate the questions needed by the machine to interact with the users.
For instance, the personal context recognition model shown in  \cite{giunchiglia2018personal}, showed to be a good approach to increase the accuracy of the context recognition algorithms. Our work described in Section~\ref{section:TheCurrentContext} enhance the representation of the personal context with the hope to perform better in real-life scenarios.
In practice, existing approaches for context recognition using batch or streaming sensor data \cite{vaizman2017context,bontempelli2020learning,zeni2019fixing} do not leverage on an explicit context modeling. The context representation is implicit in the labels used to train the machine learning model. The context formalization introduced in this work can be used to structure the output of these machine learning models according to our representation. Moreover, it can help machine learning approaches to interact with users. For instance, the machine can ask if Haonan is a friend of Xiaoyue since they are walking together.  Approaches that use active learning strategies (e.g., \cite{settles2009active,hoque2012aalo,hossain2017active}) can benefit of our representation.

Also, existing frameworks for creating context-aware mobile applications, such as~\citeauthor{ferreira2015aware}, do not consider the modelling of the context.

\section{Conclusion and Future Work}
\label{section:conclusion}

In this paper, we moved forward with a better representation of personal context in real-life environments. We proposed an improved representation of the personal context, adding the internal state, functions, and actions. The learning aspect of our work is the formal definition of an algorithm to transform the streaming input data to ML algorithms. We will put all these components in the system architecture. 

In comparison with the work on personal context recognition for human-machine collaboration \cite{giunchiglia2018personal}, we have shown an enhancement related to the model representation of the personal context.

Additionally, we have shown how our novel personal context representation can also be leveraged by machine learning algorithms to include prior knowledge about the structure of their output and can be used to drive the interaction with the user. Future work will focus on evaluating the impact of our formalization on an existing approach for fixing mislabeled data when learning the users' contexts \cite{zeni2019fixing}.

The next step is to propose and implement a modern design of the services related to iLog \cite{zeni2014multi} by a centralised streaming system and linking the personal context data collections with other distributed services of machine learning. This implementation will allow us to test and evaluate our novel context model in near real-life scenarios.

\section*{Acknowledgements}

The research conducted by Fausto and Xiaoyue has received funding from the European Union's Horizon 2020 FET Proactive project “WeNet – The Internet of us”, grant agreement No 823783.

The research conducted by Marcelo and Andrea has received funding from the \emph{“DELPhi - DiscovEring Life Patterns”} project funded by the MIUR Progetti di Ricerca di Rilevante Interesse Nazionale (PRIN) 2017 – DD n. 1062 del 31.05.2019.

\bibliographystyle{named}
\bibliography{main}

\end{document}